\pdfoutput=1
\documentclass[11pt]{article}

\usepackage[]{ACL2023}

\usepackage{times}
\usepackage{latexsym}

\usepackage{amsmath}
\usepackage{txfonts}
\usepackage{amssymb}
\usepackage{amsbsy}
\usepackage{enumitem}
\usepackage{subcaption}
\usepackage{multirow}
\usepackage{booktabs}
\usepackage{graphicx}
\usepackage{bbm}
\usepackage{makecell}
\usepackage{tablefootnote}
\usepackage{nccmath}

\usepackage[T1]{fontenc}
\usepackage[utf8]{inputenc}

\usepackage{microtype}

\usepackage{inconsolata}

\DeclareMathAlphabet\mathbfcal{OMS}{cmsy}{b}{n}

\def\LP{\Delta\mathsf{LP}}

\def\av{\mathbf{a}}
\def\bv{\mathbf{b}}
\def\bpv{\mathbf{b'}}
\def\cv{\mathbf{c}}
\def\fv{\mathbf{f}}

\def\iv{\mathbf{i}}
\def\nv{\mathbf{n}}
\def\npv{\mathbf{n'}}

\def\sv{\mathbf{s}}

\def\vv{\mathbf{v}}
\def\xv{\mathbf{x}}
\def\xpv{\mathbf{x'}}
\def\yv{\mathbf{y}}
\def\zv{\mathbf{z}}
\def\zpv{\mathbf{z'}}

\def\FM{\mathbf{F}}

\def\VM{\mathbf{V}}

\def\WM{\mathbf{W}}

\newcommand{\softmax}[2]{\underset{#1}{\text{\sc SoftMax}}(#2)}

\newcommand{\layernorm}[2]{\text{\sc N}_{#1}(#2)}
\newcommand{\layernormnoarg}[1]{\text{\sc N}_{#1}}

\newcommand{\prob}[1]{\mathsf{P}( #1 )}
\newcommand{\condprob}[2]{\mathsf{P}( #1 \mid #2 )}
\newcommand{\lp}[2]{\Delta\mathsf{LP}( #1 \mid #2 )}
\newcommand{\ff}[2]{\text{\sc FF}_{#1}(#2)}

\title{Token-wise Decomposition of Autoregressive Language Model \\ Hidden States for Analyzing Model Predictions}

\author{Byung-Doh Oh \\
  Department of Linguistics \\
  The Ohio State University \\
  \texttt{oh.531@osu.edu} \\\And
  William Schuler \\
  Department of Linguistics \\
  The Ohio State University \\
  \texttt{schuler.77@osu.edu} \\}

\begin{document}
\maketitle
\begin{abstract}
While there is much recent interest in studying why Transformer-based large language models make predictions the way they do, the complex computations performed within each layer have made their behavior somewhat opaque.
To mitigate this opacity, this work presents a linear decomposition of final hidden states from autoregressive language models based on each initial input token, which is exact for virtually all contemporary Transformer architectures.
This decomposition allows the definition of probability distributions that ablate the contribution of specific input tokens, which can be used to analyze their influence on model probabilities over a sequence of upcoming words with only one forward pass from the model.
Using the change in next-word probability as a measure of importance, this work first examines which context words make the biggest contribution to language model predictions.
Regression experiments suggest that Transformer-based language models rely primarily on collocational associations, followed by linguistic factors such as syntactic dependencies and coreference relationships in making next-word predictions.
Additionally, analyses using these measures to predict syntactic dependencies and coreferent mention spans show that collocational association and repetitions of the same token largely explain the language models' predictions on these tasks.
\end{abstract}

\section{Introduction}
\begin{figure}[t!]
    \centering
    \fbox{\includegraphics[width=0.69\linewidth]{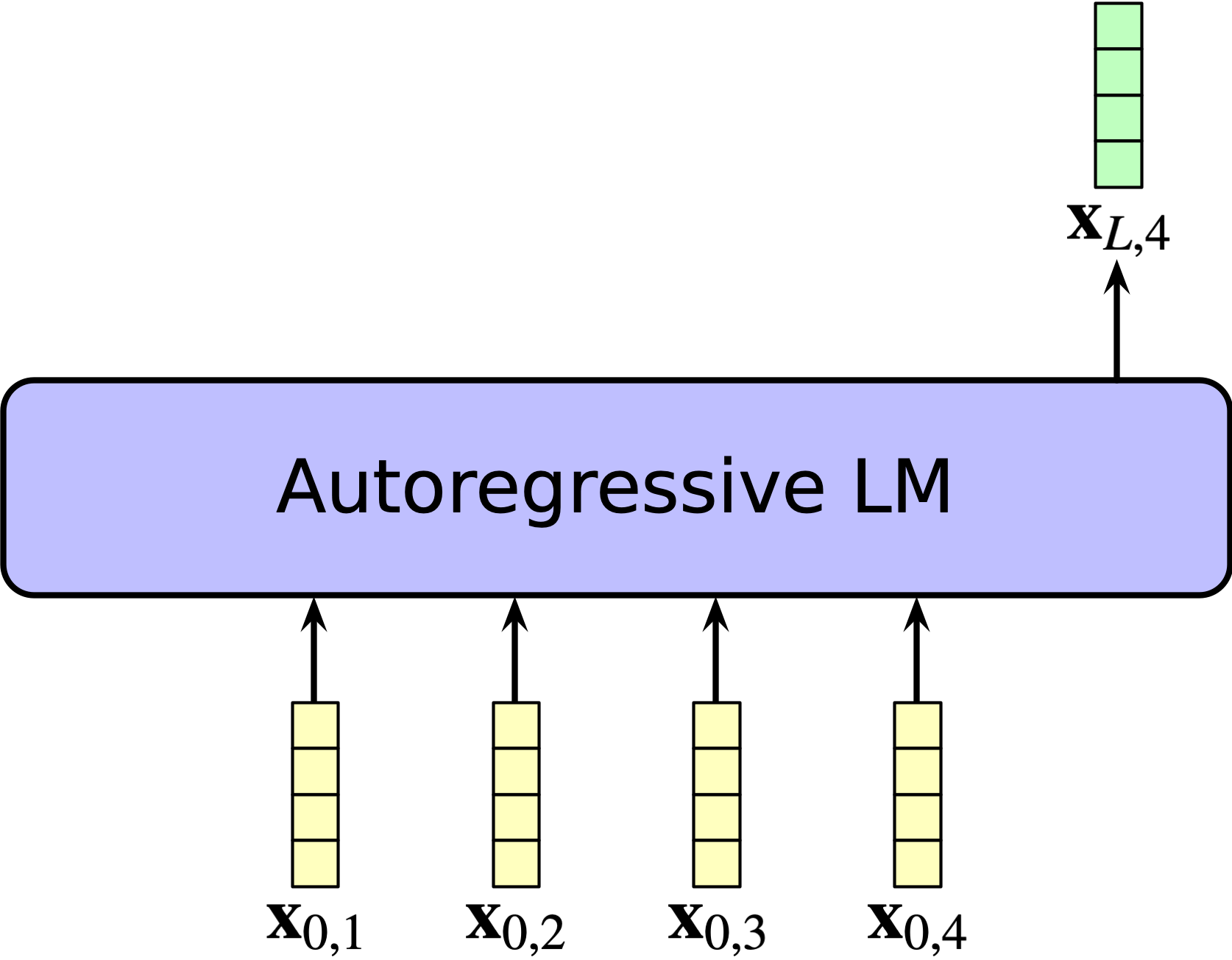}} \\[-1.5pt]
    \fbox{\includegraphics[width=0.69\linewidth]{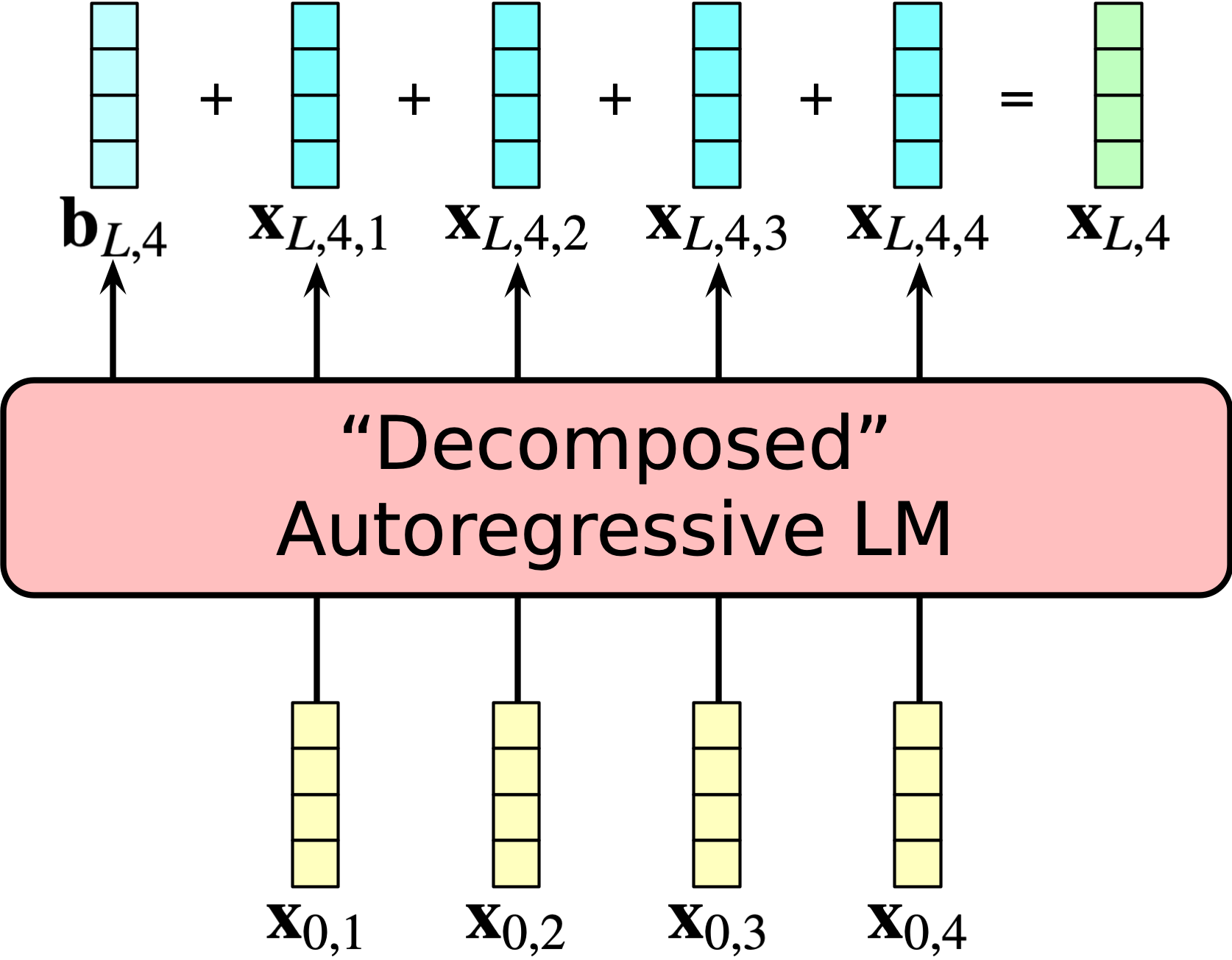}}
    \caption{Schematic of input and output representations from Transformer-based autoregressive language models. Standard models (top) calculate one vector of final hidden states at a given timestep ($\xv_{L,i}$), which in this work (bottom) is decomposed exactly into the sum of output representations of each input token ($\xv_{L,i,k}$) and a cumulative bias term ($\bv_{L,i}$).}
    \label{fig:overview}
\end{figure}
Much of contemporary natural language processing (NLP) is driven by Transformer-based large language models, which are trained to make predictions about words in their context by aggregating representations through their self-attention mechanism.
The breakthrough in many NLP tasks these models have achieved has led to active research into interpreting their predictions and probing the knowledge embodied by these models \citep{manningetal20, rogersetal21, belinkov22}. 
One line of such research focuses on quantifying the importance of each input token to the models' final output, but due to the complexity of the computations performed within the Transformer layers, analysis has been limited to studying the self-attention mechanism and the feedforward neural network independently \citep{kobayashietal20, kobayashietal21, gevaetal21, gevaetal22, mickusetal22} or has relied on e.g.~gradient-based attribution methods \citep{sanyalren21, zamanbelinkov22} that yield measures that are not interpretable in terms of output model probabilities.

To address these limitations, this work presents a linear decomposition of final language model hidden states into the sum of final output representations of each initial input token and a cumulative bias term, which is schematized in Figure \ref{fig:overview}.
This work focuses on decomposing autoregressive language models, in which the final hidden states are used to calculate a probability distribution over the next token.
The decomposition allows the definition of probability distributions that ablate the contribution of specific input tokens, which can be used to study their impact on next-token probabilities with only one forward pass from the model.
This decomposition is exact if the activation function of the feedforward neural network is differentiable almost everywhere,\footnote{That is, the function is differentiable at all real numbers except a subset of Lebesgue measure zero, such as the rectified linear unit \citep[ReLU;][]{nairhinton10}, which has an inflection point at $x=0$.} and therefore it does not require perturbing the original computations of the language model (e.g.~by using approximations) to gauge the influence of input tokens for virtually all contemporary Transformer architectures.
Additionally, this work defines an intuitive importance measure for each context token based on the change in next-token log probability, which does not correlate strongly with layer-wise attention weights or gradient norms.
Since this measure is defined in terms of log probabilities, they can also be summed to quantify importance in predicting an arbitrary sequence of tokens according to the chain rule of conditional probabilities.

Using the proposed decomposition and associated importance measure, this work characterizes which kinds of context words autoregressive language models leverage most in order to make next-word predictions.
Results from stepwise regression analyses suggest that Transformer-based language models rely mainly on collocational associations, followed by linguistic factors such as syntactic dependencies and coreference relationships.
Follow-up analyses using these importance measures to predict syntactic dependencies and coreferent mention spans additionally show that collocational association and repetitions of the same token largely explain the language models' predictions on these tasks.

\section{Background: Transformer Decoder of Autoregressive Language Models}
Transformer-based autoregressive language models \citep[e.g.][]{radfordetal19, brownetal20, zhangetal22} use a variant of the multi-layer Transformer decoder \citep{vaswanietal17transformer}.
Each decoder layer consists of a masked self-attention block and a feedforward neural network, which together calculate a vector $\xv_{l,i} \in \mathbb{R}^d$ for token $w_i$ at layer $l$:
\begin{equation}
\label{eq:decoder}
\xv_{l,i} = \ff{l}{\layernorm{l,\text{out}}{\xpv\!_{l,i} + \xv_{l-1,i}}} + (\xpv\!_{l,i} + \xv_{l-1,i}),
\end{equation}
where $\textsc{FF}_{l}$ is a two-layer feedforward neural network, N$_{l,\text{out}}$ is a vector-wise layer normalization operation, and $\xpv\!_{l,i} \in \mathbb{R}^d$ is the output representation from the multi-head self-attention mechanism, in which $H$ heads mix representations from the previous context.
This output $\xpv\!_{l,i}$ can be decomposed into the sum of representations resulting from each attention head $h$ and a bias vector~$\vv_{l}$:
\begin{equation}
\label{eq:attn}
\xpv\!_{l,i} \! = \! \sum_{h=1}^{H}\!{\VM_{l,h} \, [\layernorm{l,\text{in}}{\xv_{l-1,1}} \, \cdots \, \layernorm{l,\text{in}}{\xv_{l-1,i}}] \, \av_{l,h,i}} + \vv_{l},
\end{equation}
where $\VM_{l,h} \in \mathbb{R}^{d \times d}$ and $\vv_{l} \in \mathbb{R}^{d}$ represent the weights and biases of the composite value-output transformation\footnote{For the simplicity of notation, multi-head self-attention is formulated as a sum of `value-output' transformed representations from each attention head instead of the `output' transformed concatenation of `value' transformed representations from each attention head as in \citet{vaswanietal17transformer}. To this end, the weights and biases of the `value' and `output' transformations are respectively composed into $\VM_{l,h}$ and $\vv_{l}$. Refer to Appendix \ref{sec:value_output} for the derivation of $\VM_{l,h}$ and $\vv_{l}$.} respectively, and $\av_{l,h,i} \in \mathbb{R}^{i}$ is the vector of self-attention weights from each head.

N$_{l,\alpha}$, where $\alpha \in \{\text{in}, \text{out}\}$,\footnote{$\layernormnoarg{l,\text{in}}$ is applied before the masked self-attention block, and $\layernormnoarg{l,\text{out}}$ is applied before the feedforward neural network.} is a vector-wise layer normalization operation \citep{baetal16} that first standardizes the vector and subsequently conducts elementwise transformations using trainable parameters $\cv_{l,\alpha}, \bv_{l,\alpha} \in \mathbb{R}^{d}$:
\begin{equation}
\label{eq:ln}
\layernorm{l,\alpha}{\yv} = \frac{\yv-m(\yv)}{s(\yv)} \odot \cv_{l,\alpha} + \bv_{l,\alpha},
\end{equation}
where $m(\yv)$ and $s(\yv)$ denote the elementwise mean and standard deviation of $\yv$ respectively, and $\odot$ denotes a Hadamard product.

The output representation from the last decoder layer $L$ is layer-normalized and multiplied by the projection matrix to yield logit scores for the probability distribution over token $w_{i+1}$:
\begin{equation}
\label{eq:logits}
\zv_{i} = \WM \, \layernorm{L+1,\text{in}}{\xv_{L,i}},
\end{equation}
where $\zv_{i} \in \mathbb{R}^{V}$ is the vector of logit scores, $\WM \in \mathbb{R}^{V \times d}$ is the projection matrix, $V$ is the size of the vocabulary, and $\layernormnoarg{L+1,\text{in}}$ is the final layer normalization operation with parameters $\cv_{L+1,\text{in}}$ and $\bv_{L+1,\text{in}}$.

\section{Token-wise Decomposition of Language Model Hidden States}
This section provides a mathematical definition of the token-wise decomposition of language model hidden states, which allows the quantification of the contribution of each input token to the conditional probability of the next token.
\subsection{Mathematical Definition} \label{sec:decomposition}
In this section, we show that the vector of logits $\zv_{i}$ in Equation \ref{eq:logits} can be decomposed into the sum of final output representations of each input token $w_{k}$ and a `bias-like' term that accumulates bias vectors throughout the Transformer network, which is exact if the activation function within the feedforward neural network is differentiable almost everywhere:
\begin{equation}
    \label{eq:declogits}
    \zv_{i} = \sum\limits_{k=1}^{i}\zpv\!_{i,k} + \bv_{i},
\end{equation}
where $\zpv\!_{i,k} \in \mathbb{R}^{V}$ is the final transformed output at timestep $i$ of the input representation $\xv_{0,k}$\footnote{Throughout this paper, the input representation $\xv_{0,k}$ denotes the sum of the type-specific embedding for token $w_{k}$ and the positional embedding for position $k$.} at timestep $k$.
This $\zpv\!_{i,k}$ is calculated by aggregating the output of all computations performed on $\xv_{0,k}$ throughout the Transformer layers:
\begin{equation}
    \zpv\!_{i,k} = \WM \, \nv_{\text{x},L+1,i,k},
\end{equation}
where $\nv_{\text{x},L+1,i,k}$ is a layer-normalized version of $\xv_{L,i,k}$, explained below.
Additionally, $\mathbf{b}_{i} \in \mathbb{R}^{V}$ is the `bias-like' term resulting from accumulating computations performed on bias vectors that are difficult to attribute to any specific source position~$k$:
\begin{equation}
    % \vspace{-1mm}
    \bv_{i} = \WM \, \nv_{\text{b},L+1,i},
\end{equation}
where $\nv_{\text{b},L+1,i}$ is a layer-normalized version of $\bv_{L,i}$, also explained below.

This decomposition is in turn achieved by maintaining input-specific vectors $\xv_{l,i,k} \in \mathbb{R}^{d}$ and a `bias-like' vector $\bv_{l,i} \in \mathbb{R}^{d}$ throughout the network.
The second index of both $\xv_{l,i,k}$ and $\bv_{l,i}$ represents each target position $i$, and the third index of $\xv_{l,i,k}$ represents each source position $k \in \{1,...,i\}$.
Therefore, when the third index of $\xv_{l,i,k}$ is reduced and the result is added to $\bv_{l,i}$, the undecomposed output representation $\xv_{l,i} \in \mathbb{R}^{d}$ is returned:
\begin{equation}
    \vspace{-1mm}
    \xv_{l,i} = \sum\limits_{k=1}^{i}\xv_{l,i,k} + \bv_{l,i}.
\end{equation}
\begin{figure}[t!]
    \centering
    \includegraphics[width=\linewidth]{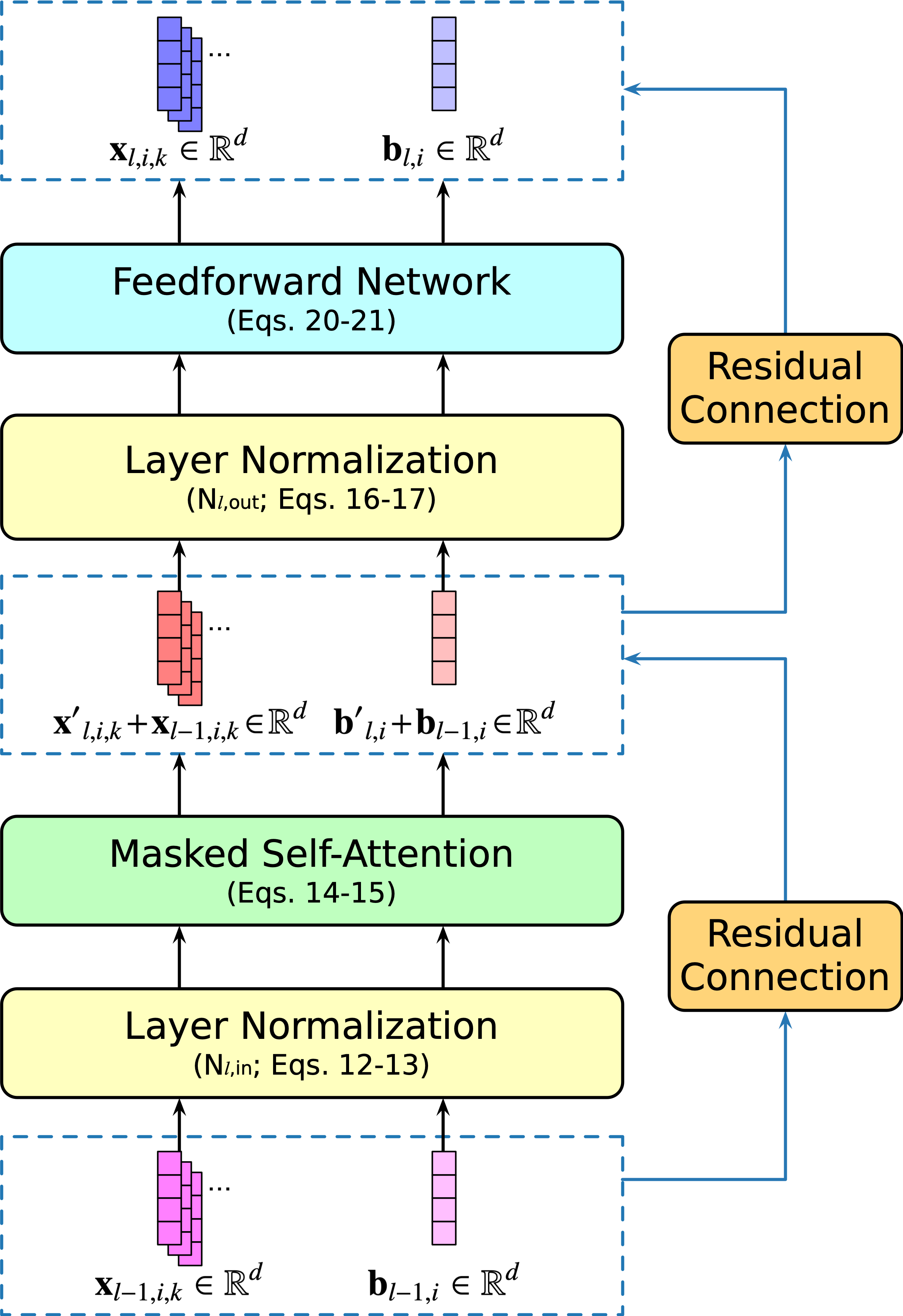}
    \caption{Alternative formulation of computations performed within one decoder layer of a Transformer-based autoregressive language model, which allows the contribution of each input token $w_k$ to $\xv_{l,i}$ to be preserved as~$\xv_{l,i,k}$.}
    \label{fig:attn_block}
\end{figure}
These decomposed representations are updated by each decoder layer (Eq.~\ref{eq:decoder}; Fig.~\ref{fig:attn_block}) as follows:
\begin{align}
\xv_{l,i,k} & = \fv_{\text{x},l,i,k} + (\xpv\!_{l,i,k} + \xv_{l-1,i,k}), \\
\bv_{l,i} & = \fv_{\text{b},l,i} + (\bpv\!_{l,i} + \bv_{l-1,i}),
\end{align}
where $\bv_{0,i} = \mathbf{0}$ and $\xv_{0,i,k}$ is a position-sensitive version of $\xv_{0,k}$:
\begin{equation}
\xv_{0,i,k} = \begin{cases} \xv_{0,k} & \text{if } i=k, \\ \mathbf{0} & \text{if } i \neq k,  \end{cases}
\end{equation}
and $\fv_{\text{x},l,i,k}$ and $\fv_{\text{b},l,i}$ are decomposed versions of the output from the feedforward network for $\xv_{l,i,k}$ and $\bv_{l,i}$, defined below.

The exact decomposition of hidden states according to each source position is made possible due to the linear nature of computations within the masked self-attention block and a local linear approximation of the activation function within the feedforward neural network.
First, layer normalization $\layernormnoarg{l,\text{in}}$ (Eq.~\ref{eq:ln}) is applied to $\xv_{l-1,i,k}$ to yield $\nv_{\text{x},l,i,k}$ by centering it, scaling it by the standard deviation of the undecomposed representation $s(\xv_{l-1,i})$, and obtaining a Hadamard product with trainable vector~$\cv_{l,\text{in}}$:
\begin{equation}
\label{eq:normstart} \nv_{\text{x},l,i,k} = \frac{\xv_{l-1,i,k}-m(\xv_{l-1,i,k})}{s(\xv_{l-1,i})} \odot \cv_{l,\text{in}}.
\end{equation}
$\layernormnoarg{l,\text{in}}$ is also applied to $\bv_{l-1,i}$ to yield $\nv_{\text{b},l,i}$, except that the bias vector $\bv_{l,\text{in}}$ is accumulated by this term:
\begin{equation}
\label{eq:normend} \nv_{\text{b},l,i} = \frac{\bv_{l-1,i}-m(\bv_{l-1,i})}{s(\xv_{l-1,i})} \odot \cv_{l,\text{in}} + \bv_{l,\text{in}}.
\end{equation}

Subsequently, the masked self-attention mechanism (Eq.~\ref{eq:attn}) is applied to $[\nv_{\text{x},l,1,k} \, \cdots \, \nv_{\text{x},l,i,k}]$ to yield $\xpv\!_{l,i,k}$, which updates the total representation from source position $k$ to target position $i$ using self-attention weights $\av_{l,h,i}$:
\begin{equation}
\label{eq:tensor_attn}
\xpv\!_{l,i,k} = \sum_{h=1}^{H}{\VM_{l,h} \, [\nv_{\text{x},l,1,k} \, \cdots \, \nv_{\text{x},l,i,k}] \, \av_{l,h,i}}.
\end{equation}
The self-attention mechanism is also applied to $[\nv_{\text{b},l,1} \, \cdots \, \nv_{\text{b},l,i}]$ to yield $\bpv\!_{l,i}$.
Similarly to layer normalization, the bias vector $\vv_{l}$ is accumulated by this term:
\begin{equation}
\bpv\!_{l,i} = \sum_{h=1}^{H}{\VM_{l,h}  \, [\nv_{\text{b},l,1} \, \cdots \, \nv_{\text{b},l,i}] \, \av_{l,h,i}} + \vv_{l}.
\end{equation}

After adding the residual representations, layer normalization $\layernormnoarg{l,\text{out}}$ is applied to $\xpv\!_{l,i,k} + \xv_{l-1,i,k}$ and $\bpv\!_{l,i} + \bv_{l-1,i}$ in a similar manner to Equations \ref{eq:normstart} and \ref{eq:normend} to yield $\npv\!_{\text{x},l,i,k}$ and $\npv\!_{\text{b},l,i}$ respectively, by centering each vector, scaling them by the standard deviation of their corresponding undecomposed representation $s(\xpv\!_{l,i}+\xv_{l-1,i})$, and applying the learned parameters $\cv_{l,\text{out}}$ and $\bv_{l,\text{out}}$:
\begin{equation}
\npv\!_{\text{x},l,i,k} = \frac{\xpv\!_{l,i,k} + \xv_{l-1,i,k}-m(\xpv\!_{l,i,k} + \xv_{l-1,i,k})}{s(\xpv\!_{l,i}+\xv_{l-1,i})} \odot \cv_{l,\text{out}},
\end{equation}
\begin{equation}
\npv\!_{\text{b},l,i} = \frac{\bpv\!_{l,i} + \bv_{l-1,i}-m(\bpv\!_{l,i} + \bv_{l-1,i})}{s(\xpv\!_{l,i}+\xv_{l-1,i})} \odot \cv_{l,\text{out}} + \bv_{l,\text{out}}.
\end{equation}

Finally, if the activation function within the feedforward neural network from Equation \ref{eq:decoder} is differentiable almost everywhere,\footnote{Virtually all widely used activation functions such as the rectified linear unit \citep[ReLU;][]{nairhinton10} and the Gaussian error linear unit \citep[GELU;][]{hendrycksgimpel16} satisfy this property.} local linear approximation can be used to calculate its output values:
\begin{align}
\label{eq:ffnn}
\ff{l}{\yv} & = \FM_{l,2} \, \sigma(\FM_{l,1} \, \yv + \fv_{l,1}) + \fv_{l,2} \\
& = \FM_{l,2}(\sv \odot (\FM_{l,1} \, \yv + \fv_{l,1}) + \iv) + \fv_{l,2},
\end{align}
where $\FM_{l,1}, \FM_{l,2}$ and $\fv_{l,1}, \fv_{l,2}$ are the weights and biases of the feedforward neural network, $\sigma$ is the activation function, and $\sv$ and $\iv$ are respectively the vector of slopes and intercepts of tangent lines specified by each element of the input vector $\FM_{l,1}\,\yv + \fv_{l,1}$.\footnote{That is, $\sv = \sigma'(\FM_{l,1}\,\yv + \fv_{l,1})$, and $\iv = \sigma(\FM_{l,1}\,\yv + \fv_{l,1}) - \sigma'(\FM_{l,1}\,\yv + \fv_{l,1}) \odot (\FM_{l,1}\,\yv + \fv_{l,1})$.}
This reformulation of the activation function allows the feedforward neural network to apply to each decomposed vector $\npv\!_{\text{x},l,i,k}$ and $\npv\!_{\text{b},l,i}$ to yield $\fv_{\text{x},l,i,k}$ and $\fv_{\text{b},l,i}$ respectively:
\begin{equation}
\fv_{\text{x},l,i,k}  = \FM_{l,2} \, \sv_{l,i} \odot \FM_{l,1}\,\npv\!_{\text{x},l,i,k}, 
\vspace{-2.5mm}
\end{equation}
\begin{equation}
\fv_{\text{b},l,i}  = \FM_{l,2}(\sv_{l,i} \odot (\FM_{l,1}\,\npv\!_{\text{b},l,i} + \fv_{l,1} ) + \iv_{l,i}) + \fv_{l,2},
\end{equation}
where $\sv_{l,i}$ and $\iv_{l,i}$ are the vector of slopes and intercepts of tangent lines specified by each element of the undecomposed $\FM_{l,1}\,\layernorm{l,\text{out}}{\xpv\!_{l,i}+\xv_{l-1,i}} + \fv_{l,1}$.  As with other operations, the bias vectors $\fv_{l,1}$, $\fv_{l,2}$, and~$\iv_{l,1}$ are accumulated by $\fv_{\text{b},l,i}$.

\subsection{Proposed Importance Measure $\LP$: Change in Next-Word Probabilities} \label{sec:lp}
Based on the decomposition outlined in Section \ref{sec:decomposition}, the importance of each input token $w_{1..i}$ to the probability of the next token $\condprob{w_{i+1}}{w_{1..i}}$ can be quantified.
To this end, the probability distribution over the next token that ablates the contribution of $w_{k}$ is defined as follows:
\begin{equation} \label{eq:ablation}
\condprob{w_{i+1}}{w_{1..i\backslash\{k\}}} = \softmax{w_{i+1}}{\zv_{i}-\zpv\!_{i,k}}.
\end{equation}
Subsequently, the importance measure of $w_k$ to the prediction of $w_{i+1}$ is calculated as the difference between log probabilities of $w_{i+1}$ given the full context ($w_{1..i}$) and the context without it ($w_{1..i\backslash\{k\}}$):
\begin{align}
& \lp{w_{i+1}}{w_{1..i},w_{k\in\{1,...,i\}}} = \\ & \log_{2}{\condprob{w_{i+1}}{w_{1..i}}} - \log_{2}{\condprob{w_{i+1}}{w_{1..i\backslash\{k\}}}}. \notag
\end{align}

This measure captures the intuition that an input token that is more crucial to predicting the next token $w_{i+1}$ will result in larger decreases in $\condprob{w_{i+1}}{w_{1..i}}$ when its contribution to the logit scores is ablated out.
It is also possible for $\LP$ to be negative, or in other words, $\condprob{w_{i+1}}{w_{1..i}}$ can increase as a result of ablating an input token $w_k$.
However, a preliminary analysis showed that negative $\LP$ values were much less commonly observed than positive $\LP$ values and input tokens with negative $\LP$ values were not in an easily interpretable relationship with the predicted token.
Therefore, the experiments in this work focus on characterizing input tokens with high $\LP$ values, which are the tokens that drive a large increase in $\condprob{w_{i+1}}{w_{1..i}}$.

\section{Experiment 1: Correlation with Other Importance Measures}
This work first compares the decomposition-based $\LP$ defined in Section \ref{sec:lp} with other measures of importance that have been used in the literature to examine the degree to which $\LP$ may be redundant with them.
To this end, Pearson correlation coefficients were calculated between the proposed $\LP$ and attention weights and gradient norms at a token level.
\begin{figure*}[ht!]
    \centering
    \includegraphics[width=\linewidth]{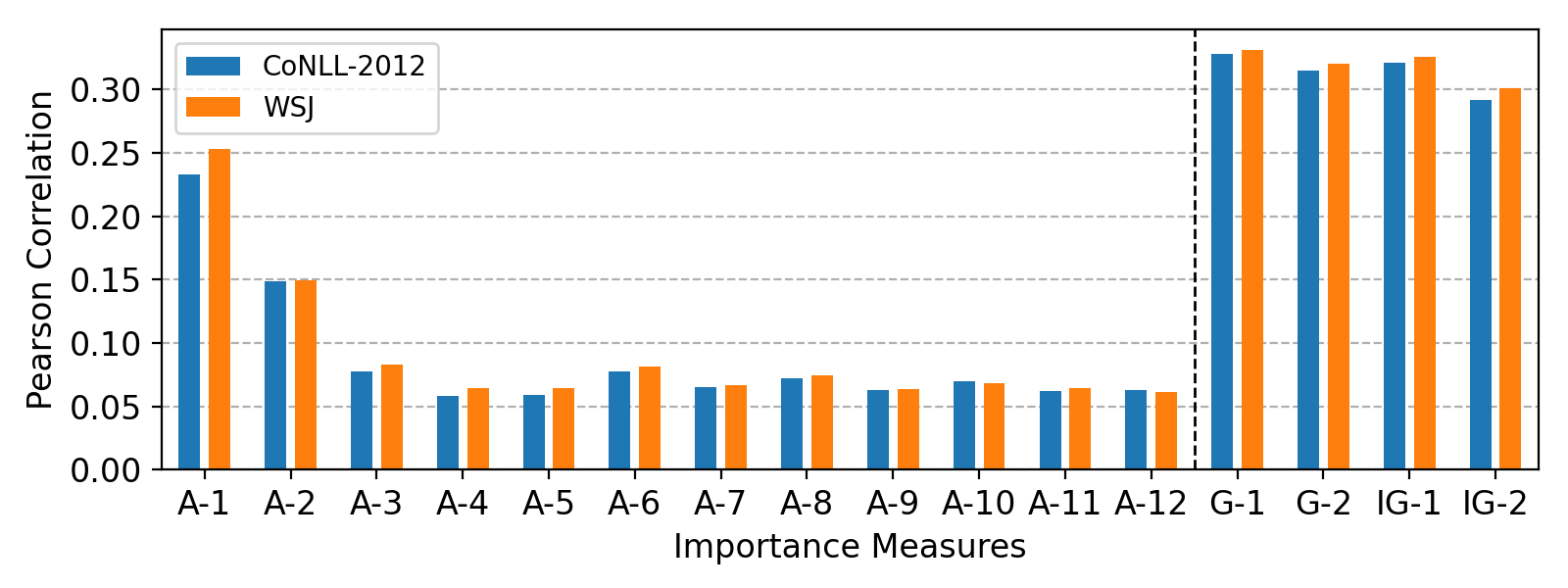}
    \caption{Pearson correlation coefficients between $\LP$ and other importance measures for each context token. A-$l$ is average attention weight at layer $l$; G-$n$ is $n$-norm of gradient; IG-$n$ is $n$-norm of input $\times$ gradient.}
    \label{fig:corr}
\end{figure*}

\subsection{Procedures}
The first experiment used the English section of the Conference on Natural Language Learning shared task corpus \citep[CoNLL-2012;][]{pradhanetal12} as well as the Wall Street Journal corpus of the Penn Treebank \citep[WSJ;][]{marcusetal93}.
Both corpora include text from the newswire domain, and the CoNLL-2012 corpus additionally includes text from broadcasts, magazines, telephone conversations, weblogs, and the Bible.
The development sets of the two corpora were used in this experiment, which consist of 9,603 and 1,700 sentences respectively.

To calculate importance measures on the two corpora, the Open Pre-trained Transformer language model \citep[OPT;][]{zhangetal22} with $\sim$125M parameters was used for efficiency.
In addition to $\LP$ defined in Section \ref{sec:lp},\footnote{Code for calculating decomposed OPT representations and their associated $\LP$ is publicly available at \url{https://github.com/byungdoh/llm_decomposition}.} the following importance measures were calculated for each context token $w_{k\in\{1,...,i\}}$ at timestep $i$:
\begin{itemize}[leftmargin=*]
    \setlength\itemsep{0em}
    \item Layer-wise attention weights \citep{vaswanietal17transformer}: Average attention weights over $w_{k}$ from all heads within each layer, i.e.~$\frac{1}{H}\sum_{h=1}^{H}\delta^{\top}_k\av_{l,h,i}$, where $\delta_{k} \in \mathbb{R}^{i}$ is a Kronecker delta vector consisting of a one at element $k$ and zeros elsewhere, and $l \in \{1,...,L\}$.
    \item Gradient norms \citep{simonyanetal14}: Norm of gradient of next-token log probability w.r.t.~the input $\xv_{0,k}$, i.e.~$||\nabla_{\xv_{0,k}}\log{\condprob{w_{i+1}}{w_{1..i}}}||_{n}$, where $n \in \{1,2\}$.
    \item Input $\times$ gradient norms \citep{shrikumaretal17}: \\  $||\xv_{0,k} \odot \nabla_{\xv_{0,k}}\log{\condprob{w_{i+1}}{w_{1..i}}}||_{n}$, where $n \in \{1,2\}$.
\end{itemize}

Each article of the CoNLL-2012 and WSJ corpora was tokenized according OPT's byte-pair encoding \citep[BPE;][]{sennrichetal15} tokenizer and was provided as input to the OPT model.
In cases where each article did not fit into a single context window, the second half of the previous context window served as the first half of a new context window to calculate importance measures for the remaining tokens.\footnote{In practice, most articles fit within one context window of 2,048 tokens.}
Finally, Pearson correlation coefficients were calculated between token-level $\LP$ and attention-/gradient-based importance measures on each corpus (163,309,857 points in CoNLL-2012; 25,900,924 points in WSJ).

\subsection{Results}

The results in Figure \ref{fig:corr} show that across both corpora, the proposed $\LP$ shows weak correlation with both attention weights and gradient norms, which suggests that $\LP$ does not capture a redundant quantity from importance measures that have been used in previous work to examine language model predictions.
The gradient norms are more correlated with $\LP$, which is likely due to the fact that the gradients calculated with respect to the original input representation $\xv_{0,k}$ accumulate all computations performed within the network like the token-wise decomposition.
However, one crucial difference between $\LP$ and gradient norms is that gradient norms can `saturate' and approach zero when the model makes accurate predictions, as $\nabla_{\zv_{i}}\log{\condprob{w_{i+1}}{w_{1..i}}} \approx \mathbf{0}$ when $\condprob{w_{i+1}}{w_{1..i}} \approx 1$. This means that the importance measures of all context tokens will be systematically underestimated for high-probability target tokens, which may be especially problematic for analyzing large language models that have been trained on billions of training tokens.
For average attention weights, they seem to correlate with $\LP$ most at layer 1, where they are calculated over layer-normalized input representations $[\layernorm{1,\text{in}}{\xv_{0,1}} \, \cdots \, \layernorm{1,\text{in}}{\xv_{0,i}}]$.
In contrast, the attention weights at higher layers seem to correlate less with $\LP$, as they are calculated over representations that have been `mixed' by the self-attention mechanism.

\section{Experiment 2: Characterizing High-Importance Context Words}
Having established that $\LP$ provides a novel method to quantify the importance of each context token to language model predictions, the second experiment conducts a series of regression analyses to characterize high-importance context words (i.e.~words with high $\LP$ values) and shed light on which kinds of context words language models leverage most in order to make predictions about the next word.

\subsection{Procedures} \label{sec:regression}
\begin{figure}[t!]
    \centering
    \includegraphics[width=\linewidth]{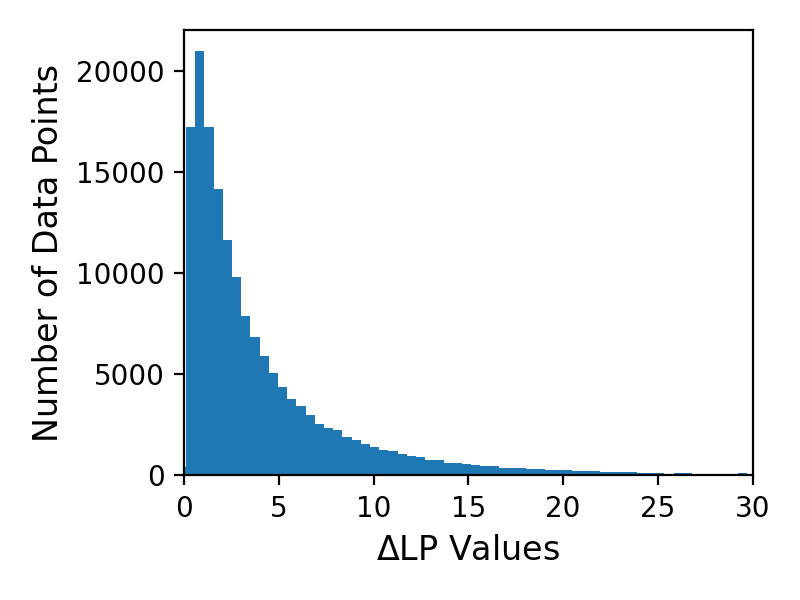}
    \caption{Histogram of top $\LP$ values for each predicted word on the development set of the CoNLL-2012 corpus calculated from the OPT model.}
    \label{fig:hist}
\end{figure}

In order to characterize high-importance context words that drive next-word predictions, linear regression models were fit in a stepwise manner to $\LP$ values on the development set of the CoNLL-2012 corpus, which contains manual annotations of both syntactic structures and coreference relationships.
To this end, the $\LP$ values were calculated for each context word at a word level (following the Penn Treebank tokenization conventions such that they align with the annotations) using the OPT model with $\sim$125M parameters.
Whenever the predicted word consisted of multiple tokens, the $\LP$ values were added together to calculate:
\begin{align} \label{eq:chainrule}
& \lp{w_{i+1}, w_{i+2}}{w_{1..i},w_{k}} = \\
& \lp{w_{i+2}}{w_{1..i+1},w_{k}} + \lp{w_{i+1}}{w_{1..i},w_{k}}, \notag
\end{align}
which is well-defined by the chain rule of conditional probabilities.
Likewise, when the context word consisted of multiple tokens, the contributions of all component tokens were ablated simultaneously (Eq.~\ref{eq:ablation}) to calculate the $\LP$ of that context word.\footnote{This ability to quantify the contribution of each context token in predicting multiple target tokens or the simultaneous contribution of multiple context tokens in model prediction is another advantage of $\LP$ over attention weights or gradient norms, which are inherently defined at a single-token level.}
In order to keep the regression models tractable, the $\LP$ value of the most important context word for each predicted word (i.e.~highest $\LP$ value) provided the response data for this experiment.
This resulted in a total of 162,882 observations, which are visualized in Figure \ref{fig:hist}.

Subsequently, a `baseline' regression model that contains baseline predictors was fit to the set of $\LP$ values.
These baseline predictors include the index of the predicted word (i.e.~how many words are in the context), the linear distance between the context word and the predicted word, and $\log{\condprob{w_{i+1}}{w_{1..i}}}$, which may be correlated with $\LP$ values.
Additionally, in order to guide the identification of factors underlying the $\LP$ values of high-importance context words, each data point was associated with the following predictors of interest that capture associations between the predicted word and the context word:
\begin{itemize}[leftmargin=*]
    \setlength\itemsep{0em}
    \item Pointwise mutual information (PMI): \\ $\log_{2}{\frac{\prob{w_{k},w_{i+1}}}{\prob{w_{k}}\prob{w_{i+1}}}}$, which is calculated using unigram and bigram probabilities estimated from the Gigaword 4 corpus \citep{parkeretal09}. Two variants of PMI are explored in this work, which capture associations of word pairs in contiguous bigrams (PMI$_{\text{bigram}}$) and document co-occurrences (PMI$_{\text{doc}}$).\footnote{The corpus was tokenized following the Penn Treebank conventions for consistency. PMI was defined to be 0 for word pairs without unigram or bigram probability estimates.}
    \item Syntactic dependency: A binary variable indicating whether the context word and the predicted word form a syntactic dependency. The CoreNLP toolkit \citep{manningetal14} was used to convert annotated constituency structures to dependency representations.
    \item Coreference relationship: A binary variable indicating whether the context word and the predicted word are in coreferent spans.
\end{itemize}

These predictors of interest were included in a stepwise manner, by including the one predictor that contributes most to regression model fit at each iteration and testing its statistical significance through a likelihood ratio test (LRT).
All predictors were centered and scaled prior to regression modeling, so the regression coefficients $\beta$ are defined in units of standard deviation and are comparable across predictors.

\subsection{Results}
The results in Table \ref{tbl:exp2} show that among the predictors of interest, both variants of PMI made the biggest contribution to regression model fit, followed by syntactic dependency and coreference relationship.\footnote{Refer to Appendix \ref{sec:delta_ll} for regression results from the first iteration of the stepwise analysis, which evaluates each predictor independently on top of the baseline regression model.}
This suggests that Transformer-based autoregressive language models rely primarily on collocational associations in making next-word predictions (e.g.~\textit{wedding} predicting \textit{groom}, \textit{medical} predicting \textit{hospital}).
Linguistic factors like syntactic dependencies and coreference relationships explained additional variance in $\LP$ values, although their contribution was not as large.

\begin{table}
\begin{center}
\begin{tabular}{l|r|r|r}
  Predictor & $\beta$ & $t$-value & $\Delta$LL \\ \hline
  Word index & 0.034 & 1.919 & - \\
  Distance & 1.126 & 62.755 & - \\
  Log prob. & -0.083 & -5.350 & - \\ \hline
  PMI$_\text{bigram}$ & 1.220 & 70.857 & 6151.262$^{*}$ \\
  PMI$_\text{doc}$ & 1.286 & 73.952 & 3194.815$^{*}$  \\
  Dependency & 1.055 & 63.720 & 1981.778$^{*}$  \\
  Coreference & 0.123 & 7.195 & 25.883$^{*}$  \\
\end{tabular}
\end{center}
\caption{Regression coefficients from the final stepwise regression model and increase in regression model likelihood ($\Delta$LL) from including each predictor of interest. The predictors of interest are presented in the order they were included during stepwise regression (i.e.~strongest predictor at each iteration). *: $p<0.001$.}
\label{tbl:exp2}
\end{table}

The baseline predictors also shed light on the characteristics of context words that have a large influence on next-word probabilities.
Most notably, the linear distance between the predicted word and the context word was a positive predictor of $\LP$, which indicates that language models can leverage words far back in the context and that the contribution of such context words is large when they do.
Moreover, $\LP$ values were negatively correlated with log probability, which indicates that the contribution of context words generally decreases when the model is making confident predictions about the next word.
Finally, although there was a positive correlation between word index and $\LP$ values, its strength was too weak to draw conclusive interpretations.

\section{Experiment 3: Syntactic Dependency and Coreference Prediction Using $\LP$}
The previous experiment revealed that compared to measures of collocational association, syntactic dependency and coreference relationships were not as strong predictors of $\LP$.
Experiment 3 further examines the connection between high-importance context words and syntactic dependency and coreference relationships by using $\LP$ to predict them independently and analyzing the extent to which each relationship type aligns with $\LP$.

\subsection{Procedures}
This experiment used $\LP$ to make predictions about context words in syntactic dependency and coreference relationships on the development sets of the WSJ and CoNLL-2012 corpora respectively.

First, on the WSJ corpus, the precision scores for syntactic dependency relations were calculated by counting how many times context words with high $\LP$ match words in syntactic dependency relations.
While each word has exactly one incoming typed edge from its head in a typical dependency syntax representation, since autoregressive language models have no access to the forward context, all edges between word pairs were treated as undirected edges and were evaluated at the later word in the pair.
For each predicted word $w_{i+1}$ that is in $n$ syntactic dependency relationships, the top-$n$ context words were selected based on $\LP$ within the same sentence and compared to the $n$ words that are in syntactic dependency relationships with $w_{i+1}$.
The syntactic dependency representations converted using the CoreNLP toolkit \citep{manningetal14} were used to evaluate the performance on the WSJ corpus.
As a baseline, the expected precision scores from randomly selecting $n$ previous words within the same sentence are also reported.

Similarly, antecedent selection precision scores for coreference relations were calculated by counting how many times the context word with the highest $\LP$ value matched words in spans denoting the same entity.
For each mention span, $\LP$ quantifying the impact of every context word on the prediction of the entire span (Eq.~\ref{eq:chainrule}) was calculated.
Subsequently, the context word with the highest $\LP$ was evaluated in terms of whether it belonged to any antecedent spans denoting the same entity.
As a baseline, precision scores from selecting the most recent word with the same part-of-speech as the head word of the span are reported.

\begin{table}
\begin{center}
\begin{tabular}{l|r|r||r|r}
  Relation & $\LP$ & Base. & PMI$_\text{b}$ & PMI$_\text{d}$ \\ \hline
  Nom.~subj. & 61.15 & 39.79 & 1.38 & 1.44 \\ % nsubj
  Direct obj. & 70.43 & 22.01 & 0.91 & 1.57 \\ % obj
  Oblique & 52.54 & 24.31 & -0.68 & 1.54 \\ % obl
  Compound & 80.44 & 39.56 & 4.97 & 2.93 \\ % compound
  Nom.~mod. & 53.84 & 26.09 & -0.41 & 1.84 \\ % nmod
  Adj.~mod. & 82.55 & 36.02 & 4.36 & 2.17 \\ % amod
  Determiner & 52.03 & 36.52 & 1.51 & 1.08 \\ % det
  Case marker & 52.38 & 27.96 & -0.29 & 1.08 \\ \hline % case
  Microavg. & 56.20 & 29.22 & 1.11 & 1.58
\end{tabular}
\end{center}
\caption{Precision scores calculated using $\LP$, random word baseline, and average PMI of frequent syntactic dependency relations in the WSJ corpus. The less frequent relations are not presented separately but are included in the microaverage. PMI$_\text{b}$ is average PMI based on contiguous bigrams; PMI$_\text{d}$ is average PMI based on document co-occurrences.}
\label{tbl:exp3dep}
\end{table}

\subsection{Results}
The syntactic dependency results in Table \ref{tbl:exp3dep} reveal a discrepancy in performance according to the type of relation that is being predicted.
Generally, context words with high $\LP$ values corresponded most closely to words in adjectival modifier and compound relations, followed by those in subject and direct object relations, which are core arguments in English.
Performance on adjunct nouns such as nominal modifiers and oblique nouns, as well as function words like determiners and case markers was lower.
This trend in turn seems to be generally driven by the strength of collocational associations, as can be seen by the corresponding average PMI values in Table \ref{tbl:exp3dep}.
This corroborates the regression results of Experiment 2 and further suggests that the seeming connection between language model predictions and syntactic dependencies may underlyingly be the effects of collocational association.
One counterexample to this trend seems to be the syntactic dependency between the main verb and its direct object, which shows close correspondence to $\LP$ despite not having high average PMI values.

\begin{table}[t!]
\begin{center}
\begin{tabular}{l|r|r||r}
  Mention head POS & $\LP$ & Base. & Rep.\% \\ \hline
  Personal pronoun & 26.55 & 36.80 & 30.92 \\ % PRP
  Possessive pronoun & 23.29 & 36.45 & 30.59 \\ % PRP$
  Proper noun (sg.) & 61.21 & 23.19 & 68.80 \\ % NNP
  Proper noun (pl.) & 70.67 & 57.33 & 68.00 \\ % NNPS
  Common noun (sg.) & 43.39 & 12.55 & 48.75 \\ % NN
  Common noun (pl.) & 47.01 & 24.73 & 55.03 \\ % NNS
  Possessive ending & 46.28 & 30.58 & 40.91 \\ \hline % POS
  Microavg. & 38.21 & 28.65 & 43.26
\end{tabular}
\end{center}
\caption{Precision scores calculated using $\LP$, most recent head POS baseline, and Rep.~\% of frequent types of coreferent spans in the CoNLL-2012 corpus. The less frequent types are not presented separately but are included in the microaverage. Rep.~\% is the proportion of mention spans whose head words are repeated from previous coreferent spans.}
\label{tbl:exp3coref}
\end{table}

The coreference results in Table \ref{tbl:exp3coref} show an even larger gap in performance according to the type of entity mention.
Generally, context words with high $\LP$ values corresponded most closely to previous mentions of proper nouns and common nouns.
In contrast, they did not correspond well to antecedents of personal and possessive pronouns, showing lower precision scores than a simple baseline that chooses the most recent pronoun.
A follow-up analysis of the $\LP$ values showed that when the language model has to predict a head word that has already been observed in its context, the earlier occurrence of that head word contributes substantially to its prediction.
The proportion of mention spans whose head words are repeated from head words of previous coreferent spans in Table \ref{tbl:exp3coref} shows that the close correspondence between $\LP$ and previous mentions of proper nouns is driven by the fact that these proper nouns are often repeated verbatim in the corpus.
In contrast, the prediction of pronouns does not seem to be mainly driven by context words that denote their antecedents.

\section{Discussion and Conclusion}
This work advances recent efforts to interpret the predictions of Transformer-based large language models.
To this end, a linear decomposition of final language model hidden states into the sum of final output representations of each initial input token and a cumulative bias term was presented.
This decomposition is exact as long as the activation function of the feedforward neural network is differentiable almost everywhere, and therefore it is applicable to virtually all Transformer-based architectures.
Additionally, this decomposition does not require perturbing any intermediate computations nor re-running the language model to examine the impact of each input token.
The decomposition in turn allows the definition of probability distributions that ablate the influence of input tokens, which was used to define the importance measure $\LP$ that quantifies the change in next-token log probability.
The first experiment in this work demonstrated that $\LP$ does not capture a redundant quantity from importance measures that have been used in previous work to examine language model predictions such as layer-wise attention weights or gradient norms.

Subsequently, based on the proposed $\LP$, a stepwise regression analysis was conducted to shed light on the characteristics of context words that autoregressive language models rely on most in order to make next-word predictions.
The regression results show that Transformer-based language models mainly leverage context words that form strong collocational associations with the predicted word, followed by context words that are in syntactic dependencies and coreference relationships with the predicted word.
The high reliance on collocational associations is consistent with the mathematical analysis of Transformers that a layer of self-attention effectively functions as a lookup table that tracks bigram statistics of the input data \citep{elhageetal21}, as well as empirical observations that Transformer-based autoregressive language models have a propensity to `memorize' sequences from the training data \citep{carlinietal22}.

Finally, as a follow-up analysis, $\LP$ was used to predict syntactic dependencies and coreferent mentions to further examine their relationship to high-importance context words.
The precision scores on both tasks revealed a large discrepancy in performance according to the type of syntactic dependency relations and entity mentions.
On syntactic dependency prediction, $\LP$ corresponded closer to words in relations with high collocational association such as compounds and adjectival modifiers, providing further support for its importance in a language model's next-word prediction.
Moreover, on coreferent antecedent selection, $\LP$ more accurately identified previous mentions of proper nouns and common nouns that were already observed verbatim in context.
This is consistent with the tendency of Transformer-based language models to predict identical tokens from its context \citep{sunetal21}, which seems to be enabled by dedicated `induction heads' \citep{elhageetal21, olssonetal22} that learn such in-context copying behavior.

Taken together, these results suggest that collocational association and verbatim repetitions strongly drive the predictions of Transformer-based autoregressive language models.
As such, the connection drawn between a large language model's computations and linguistic phenomena such as syntactic dependencies and coreference observed in previous work \citep[e.g.][]{manningetal20} may underlyingly be the effects of these factors.

\section*{Acknowledgments}
We thank the reviewers for their helpful comments.
This work was supported by the National Science Foundation grant \#1816891.
All views expressed are those of the authors and do not necessarily reflect the views of the National Science Foundation.

\section*{Limitations}
The connection between factors underlying the predictions of Transformer-based autoregressive language models and linguistic factors drawn in this work is based on a model trained on English text and annotated corpora of English text.
Therefore, this connection may not generalize to other languages with e.g.~more flexible word order.
Additionally, although the alternative formulations of Transformer hidden states yielded insights about language model predictions, they are more computationally expensive to calculate as they rely on an explicit decomposition of the matrix multiplication operation, which in undecomposed form is highly optimized for in most packages.

\section*{Ethics Statement}
Experiments presented in this work used datasets from previously published research \citep{pradhanetal12, marcusetal93}, in which the procedures for data collection, validation, and cleaning are outlined.
These datasets were used to study a large language model's predictions about coreference resolution and dependency parsing respectively, which is consistent with their intended use.
As this work focuses on studying the factors underlying the predictions of large language models, its potential risks and negative impacts on society seem to be minimal.

\bibliography{acl2023}
\bibliographystyle{acl_natbib}

\appendix
\section{Composition of `Value' and `Output' Transformations}
\label{sec:value_output}
In \citeauthor{vaswanietal17transformer}'s \citeyearpar{vaswanietal17transformer} formulation of multi-head attention, the `value' transformation is defined at the head level with weights $\WM^\mathrm{V}_{l,h} \in \mathbb{R}^{(d/H) \times d}$ and biases $\bv^\mathrm{V}_{l,h} \in \mathbb{R}^{(d/H)}$, and the `output' transformation is defined at the layer level with weights $\WM^\mathrm{O}_{l} \in \mathbb{R}^{d \times d}$ and biases $\bv^\mathrm{O}_{l} \in \mathbb{R}^{d}$.
$\VM_{l,h}$ and $\vv_{l}$ defined in Equation \ref{eq:attn} are equal to:
\begin{align}
\VM_{l,h} & = \WM^\mathrm{O}_{l} (\delta_{h} \otimes \WM^\mathrm{V}_{l,h}), \\
\vv_{l} & = \sum_{h=1}^{H}{\WM^\mathrm{O}_{l} (\delta_{h} \otimes \bv^\mathrm{V}_{l,h}}) + \bv^\mathrm{O}_{l},
\end{align}
where $\delta_{h} \in \mathbb{R}^{H}$ is a Kronecker delta vector consisting of a one at element $h$ and zeros elsewhere, and $\otimes$ denotes a Kronecker product.

\section{Additional Regression Results}
\label{sec:delta_ll}

Regression results from the first iteration of the stepwise analysis in Experiment 2, which evaluates each predictor of interest independently on top of the baseline regression model, are outlined in Table~\ref{tbl:exp2_first}.
%
% baseline: -540192.787989695
% pmi_b: -534041.526459562
% pmi_d: -535117.247077668
% dep: -536332.934352509
% coref: -539953.840231918
% pmi_b + pmi_d: -530846.711037315
% pmi_b + pmi_d + dep: -528864.933118069
% pmi_b + pmi_d + dep + coref: -528839.050430409
%
\begin{table}[ht!]
\begin{center}
\begin{tabular}{l|r|r|r}
  Predictor & $\beta$ & $t$-value & $\Delta$LL \\ \hline
  PMI$_\text{bigram}$ & 1.832 & 113.043 & 6151.262$^{*}$ \\
  PMI$_\text{doc}$ & 1.643 & 102.341 & 5075.541$^{*}$  \\
  Dependency & 1.462 & 88.912 & 3859.854$^{*}$  \\
  Coreference & 0.362 & 21.877 & 238.948$^{*}$  \\
\end{tabular}
\end{center}
\caption{Regression coefficients and increase in regression model likelihood ($\Delta$LL) from regression models that include one predictor of interest on top of the baseline regression model. *: $p<0.001$.}
\label{tbl:exp2_first}
\end{table}

\end{document}